\documentclass[journal,twoside,web]{ieeecolor}
\usepackage{generic}
\usepackage{cite}
\usepackage{amsmath,amssymb,amsfonts}
\usepackage{algorithmic}
\usepackage{graphicx}
\usepackage{algorithm,algorithmic}
\usepackage{hyperref}
\usepackage{textcomp}
\usepackage{amsmath}
\usepackage{amssymb}
\usepackage{colortbl}
\usepackage{booktabs}
\def\BibTeX{{\rm B\kern-.05em{\sc i\kern-.025em b}\kern-.08em
    T\kern-.1667em\lower.7ex\hbox{E}\kern-.125emX}}
\newcommand{\smat}[1]{{\scriptsize #1}}
\begin{document}
\title{P2E-VQ: ECG-linked representation augmentation
for PPG via discrete patch retrieval}
\author{
Zhongli Wu\textsuperscript{$\dagger$},
Zhuangzhi Gao\textsuperscript{$\dagger$},
He Zhao,
Feixiang Zhou,
Fu Wang,
Jinru Ding,
Yuankai Wang,
Hongyi Qin,
Gregory Y. H. Lip,
Bil Kirmani,
and Yalin Zheng
\thanks{\textsuperscript{$\dagger$}These authors contributed equally to this work.}
\thanks{Zhongli Wu, Yuankai Wang, and Gregory Y. H. Lip are with the Liverpool Centre for Cardiovascular Science, University of Liverpool, Liverpool, United Kingdom.}
\thanks{Yalin Zheng, He Zhao, Feixiang Zhou, Hongyi Qin and Fu Wang are with the Department of Eye and Vision Sciences, University of Liverpool, Liverpool, United Kingdom.}
\thanks{Zhuangzhi Gao and Jinru Ding are with the Shanghai Artificial Intelligence Laboratory, Shanghai, China.}
\thanks{Bil Kirmani is with the Department of Cardiothoracic Surgery, Liverpool Heart \& Chest Hospital NHS Trust, Liverpool, United Kingdom.}
\thanks{\raggedright Corresponding author: Yalin~Zheng (e-mail: \href{mailto:Yalin.Zheng@liverpool.ac.uk}{Yalin.Zheng@liverpool.ac.uk}).}
}
\maketitle

\begin{abstract}

Photoplethysmography (PPG) is widely used in consumer wearables because of its low cost and ease of acquisition. However, unlike electrocardiography (ECG), PPG measures peripheral pulse dynamics rather than cardiac electrical activity, limiting its ability to predict cardiac conditions that rely on ECG-specific morphological cues. Existing methods attempt to bridge this gap by reconstructing ECG from PPG, but this inverse mapping is inherently ill-posed, and faithful waveform reconstruction does not necessarily translate into better downstream performance. To address this challenge, we propose P2E-VQ, a retrieval-augmented framework that replaces ECG waveform reconstruction with ECG-linked representation retrieval. Specifically, P2E-VQ converts PPG patches into discrete tokens and retrieves ECG-linked information from a memory bank constructed exclusively from the training data, thereby augmenting PPG representations while requiring only PPG during inference. Extensive experiments on five public datasets covering six downstream tasks, including clinical endpoint prediction and affective-state recognition, demonstrate that P2E-VQ consistently improves over pretrained baselines under a unified frozen-feature linear-probing protocol.

\end{abstract}

\begin{IEEEkeywords}
Photoplethysmography, Electrocardiography, Vector Quantization, Retrieval-Augmented Learning, Wearable Health Monitoring
\end{IEEEkeywords}

\section{Introduction}
\label{sec:introduction}
\IEEEPARstart{P}{hotoplethysmography} (PPG) has become the core sensing modality in consumer wearables, such as smartwatches, fitness bands, and ring-type devices, enabling scalable and passive monitoring in real-world settings \cite{charlton_2023_2023, castaneda2018review}. PPG measures pulse-induced blood-volume changes at peripheral sites using optical sensors, thereby capturing information related to heart rate dynamics, vascular properties, and cardiovascular status \cite{allen2007photoplethysmography}. Several PPG-derived indices, such as pulse-rate variability, have been shown to correlate with ECG-based heart-rate variability under many conditions, motivating efforts to extract richer cardiac information from PPG beyond simple rate tracking \cite{schafer_how_2013,gil_photoplethysmography_2010}. Nevertheless, PPG remains an indirect hemodynamic surrogate and lacks the electrophysiological morphology that makes ECG diagnostically informative, particularly for rhythm and conduction abnormalities \cite{allen2007photoplethysmography,castaneda2018review}.

Electrocardiography (ECG) provides direct measurements of cardiac electrical activity and remains the clinical reference standard for diagnosing rhythm disturbances, conduction abnormalities and a broad range of cardiovascular conditions \cite{kligfield_ecg_standard_2007}. In standard 12-lead recordings, limb lead~II is routinely used for rhythm monitoring because its electrical axis aligns closely with the main depolarization vector, yielding a clear and stable P--QRS--T morphology \cite{dubin2000rapid}. Despite its diagnostic value, ECG is less suitable for scalable and continuous real-world monitoring because it typically requires electrode contact and is often limited to short, user-initiated recordings.

Given this gap, a natural question is \textbf{\textit{whether ECG-linked electrophysiological information can be recovered or transferred from PPG signals}}. This question has motivated a growing body of PPG-to-ECG studies, which aim to learn waveform-level correspondences between paired PPG and ECG recordings. Early PPG-to-ECG methods learned waveform-level mappings using regression-based or transform-domain approaches \cite{zhu2019ecg,tian2020crossdomain}; however, their point-wise reconstruction losses are sensitive to temporal misalignment and often produce over-smoothed average ECG morphologies. More recently, deep neural networks and generative models have been introduced to synthesize ECG waveforms from PPG signals \cite{sarkar2021cardiogan,belhasin_uncertainty-aware_2025,nambu_cardioflow_2025,vo2021p2ewgan}; however, fully parametric generation may hallucinate ECG patterns that are not physiologically faithful to the individual subject. Critically, PPG does not uniquely determine ECG morphology: similar PPG pulse patterns may correspond to different ECG P--QRS--T morphologies across individuals. Therefore, subject-specific ECG reconstruction is inherently ambiguous, and reconstructed ECG waveforms may provide unreliable features for downstream prediction.

To address this challenge, we propose P2E-VQ (PPG-to-ECG with Vector Quantization),  a retrieval-augmented framework that bypasses subject-specific ECG waveform reconstruction by using discrete PPG tokens to retrieve ECG-linked information and augment PPG representations. Throughout this paper, we use \emph{ECG-linked} to refer to information carried by real ECG patches that are paired with PPG in the training data and retrieved at inference; it denotes ECG-derived evidence transferred into the PPG representation, not a reconstruction of the subject's own ECG. Specifically, during paired-data pretraining, P2E-VQ clusters short PPG patches into discrete tokens and associates each token with a set of temporally aligned ECG patches. Given a PPG-only recording, each patch is assigned to a token, the corresponding ECG-linked candidates are retrieved and aggregated, and the resulting ECG-linked representation is combined with the original PPG representation for downstream prediction. This token-indexed memory also makes retrieval efficient: ECG candidates are fetched directly from the matched memory entry rather than searched over the entire training set. Crucially, by operating at the representation level rather than performing deterministic PPG-to-ECG inversion, P2E-VQ does not require a single PPG pattern to map to a unique ECG waveform. Retrieved and aggregated ECG-linked candidates serve as evidence-level cues, yielding more discriminative features for tasks where ECG morphology provides complementary information beyond PPG alone. The main contributions of this work are summarized as follows:

\begin{itemize}
\item We reformulate PPG-to-ECG learning from waveform reconstruction to retrieval-augmented representation learning. Because the PPG-to-ECG mapping is inherently ill-posed, we bypass subject-specific ECG estimation and instead transfer ECG-linked electrophysiological information into PPG representations for downstream prediction.

\item We design a P2E-VQ mechanism that discretizes local PPG morphologies into codebook prototypes and links each prototype to aligned ECG patches, enabling non-parametric ECG-linked retrieval instead of direct waveform generation.

\item We introduce an efficient token-indexed retrieval (ETIR) strategy, where ECG candidates are organized by their associated PPG tokens and retrieved from the matched memory entry, avoiding exhaustive search over the full training set.

\item We demonstrate consistent improvements over strong pretrained baselines across five public datasets and six downstream tasks, with matched-capacity negative controls confirming that the gains are attributable to ECG-linked content rather than feature dimensionality.

\end{itemize}

\section{RELATED WORK}
\subsection{PPG-to-ECG Cross-Modal Translation}
PPG-to-ECG cross-modal translation has been widely studied as an ECG waveform reconstruction problem, aiming to recover ECG-linked information from PPG signals. Traditional methods typically learn direct waveform mappings using handcrafted features, regression-based models, transform-domain representations, dictionary learning, or kernel-based approaches \cite{banerjee2014photoecg,zhu2019ecg,tian2020crossdomain,ho2022quickly}. For example, PhotoECG predicts ECG interval ranges from handcrafted PPG features \cite{banerjee2014photoecg}, while Zhu et al. map discrete cosine transform (DCT) coefficients of PPG beats to those of corresponding ECG beats \cite{zhu2019ecg}. Although these methods can produce visually plausible ECG reconstructions, direct waveform reconstruction is sensitive to temporal misalignment and may yield over-smoothed or physiologically unfaithful ECG morphologies.

With the rise of deep learning, recent PPG-to-ECG studies have moved from handcrafted mappings to fully data-driven nonlinear reconstruction models. Recurrent and convolutional architectures, such as subject-specific BiLSTM models and end-to-end PPG2ECGps, have been used to reconstruct single-lead ECG signals from PPG recordings \cite{tang2022robust,tang2023ppg2ecgps,ezzat2024hybrid}. Generative models further improve waveform realism; for example, CardioGAN uses adversarial learning to synthesize ECG from PPG, while diffusion-based models such as RDDM generate high-fidelity ECG waveforms by modeling region-specific ECG structures \cite{sarkar2021cardiogan,shome_region-disentangled_2023,li2024biodiffusion}. Although these approaches can improve visual fidelity, they still rely on fully parametric waveform generation and may produce over-smoothed or hallucinated ECG patterns that are not physiologically faithful to individual subjects, limiting their reliability for downstream prediction.

Despite these advances, reconstruction-based methods share a common limitation: they implicitly assume that ECG morphology can be reliably inferred from PPG. However, similar peripheral pulse patterns may correspond to different P--QRS--T morphologies across individuals, making subject-specific ECG reconstruction inherently ambiguous. In contrast, P2E-VQ avoids deterministic waveform inversion by retrieving and aggregating ECG-linked patch candidates from paired training data, using them as representation-level evidence to enrich PPG features.

\subsection{Retrieval-Augmented Learning}

Retrieval-augmented learning supplements parametric model knowledge with information retrieved from an external non-parametric memory at inference time \cite{lewis2020rag, khandelwal2021nearestneighbormachinetranslation}, and has demonstrated strong performance in natural language processing and medical image analysis. Vector quantization provides a natural bridge between continuous representations and discrete retrieval keys: VQ-VAE \cite{oord2017neural} replaces continuous latent variables with codebook indices, forcing the encoder to map inputs onto a finite set of prototype vectors and enabling each prototype to serve as an efficient key for memory lookup. 

In biomedical signal analysis, vector-quantized representations have long been used for ECG compression, where codebooks capture recurring beat morphologies such as QRS complexes and ST-segment deviations \cite{sun2005beatbased}. Self-supervised pre-training has likewise been shown to yield ECG representations that transfer across datasets and tasks \cite{sarkar2022selfsupervised}. ECG foundation models further show that large unlabeled ECG corpora can support transferable latent representations for downstream tasks \cite{zhang2025ecgfm}. However, existing quantization-based methods typically use discrete codes as internal compression variables or pretext targets, with the learned codes ultimately consumed by a parametric decoder or classifier rather than used for explicit cross-modal retrieval.

In contrast, P2E-VQ applies retrieval-augmented vector quantization to cross-modal physiological representation learning. Rather than reconstructing subject-specific ECG waveforms, P2E-VQ uses PPG-derived discrete tokens to retrieve ECG-linked information from paired training data and integrate it into PPG representations. This design enables ECG-linked inference from PPG-only recordings without parametric waveform generation, combining the accessibility of PPG with the electrophysiological informativeness of ECG.

\section{Methodology}

\begin{figure*}[ht]
\begin{center}
\includegraphics[width=0.9\linewidth,height=11.5cm]{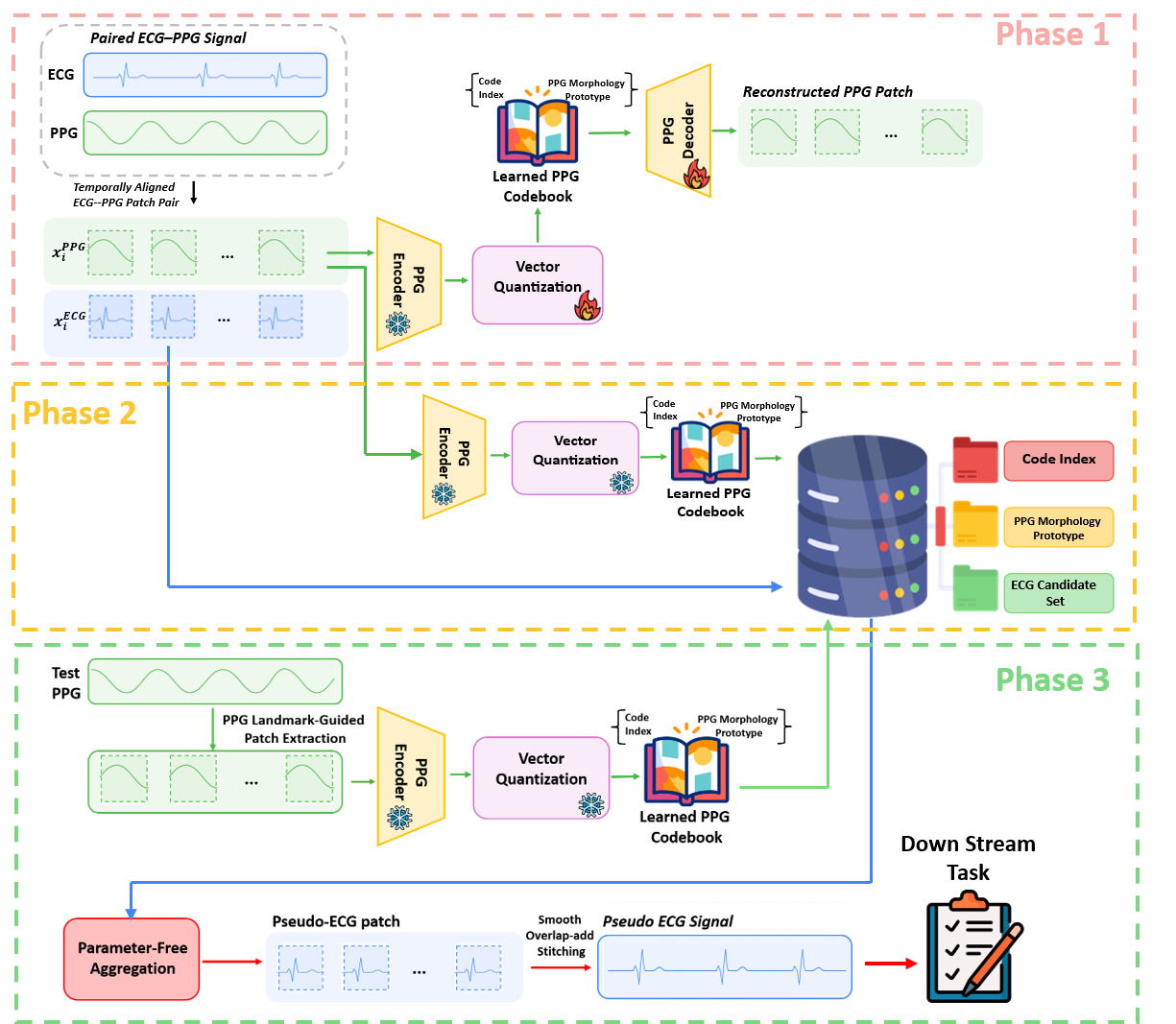}
\end{center}
\caption{Overview of P2E-VQ, a three-phase framework for ECG-linked PPG representation learning. Phase 1 learns discrete ECG-aligned PPG tokens from paired PPG--ECG patches using a vector-quantized codebook. Phase 2 uses these tokens to index temporally aligned ECG patches and construct an ECG patch memory. Phase 3 tokenizes unseen PPG-only recordings, retrieves ECG-linked patches from memory, and integrates the retrieved information with PPG features to form the final ECG-linked representation.}
\label{fig:framework}
\end{figure*}

\subsection{Overall Framework} 
As shown in Fig.~\ref{fig:framework}, P2E-VQ is a three-phase retrieval-augmented framework for ECG-linked PPG representation learning. In Phase 1, paired PPG--ECG recordings are segmented into aligned patches, and a vector-quantized codebook is learned to convert local PPG morphologies into discrete tokens. In Phase 2, these tokens are used to index temporally aligned ECG patches from the training data, forming a token-indexed ECG memory. This memory implements the ETIR strategy,  where ECG candidates are grouped by their associated PPG tokens and can be fetched from the matched memory entry rather than searched over the entire training set. In Phase 3, a PPG-only recording is tokenized using the learned codebook, and the corresponding ECG-linked candidates are retrieved and aggregated from memory. The retrieved information is then integrated with PPG features to form the final ECG-linked representation for downstream prediction. This design enables ECG-linked inference from PPG-only signals while avoiding subject-specific ECG waveform reconstruction.

\subsection{Phase 1: PPG Codebook Learning} 

The input to Phase 1 is a synchronized ECG--PPG signal pair. 
Through temporally aligned patch extraction, we obtain ECG--PPG 
patch pairs $\{(x_i^{\mathrm{ECG}}, x_i^{\mathrm{PPG}})\}_{i=1}^{N}$, 
where $x_i^{\mathrm{ECG}}, x_i^{\mathrm{PPG}} \in \mathbb{R}^{t_{\mathrm{len}}}$ 
denote ECG and PPG patches of length $t_{\mathrm{len}}$, respectively. 
The goal of this phase is to learn a discrete codebook for local 
PPG morphology. Each PPG patch $x_i^{\mathrm{PPG}}$ is encoded, assigned to its nearest code vector, and decoded to reconstruct 
the original PPG patch. This reconstruction objective encourages 
the codebook to capture recurring local pulse-shape patterns in PPG. 
The outputs of Phase 1 are a trained PPG codebook 
$\mathcal{C}$ and token assignments $\{q(i)\}_{i=1}^{N}$, 
where $q(i)\in\{1,\ldots,K\}$.

\begin{figure}[t]
\centering
\includegraphics[width=\columnwidth]{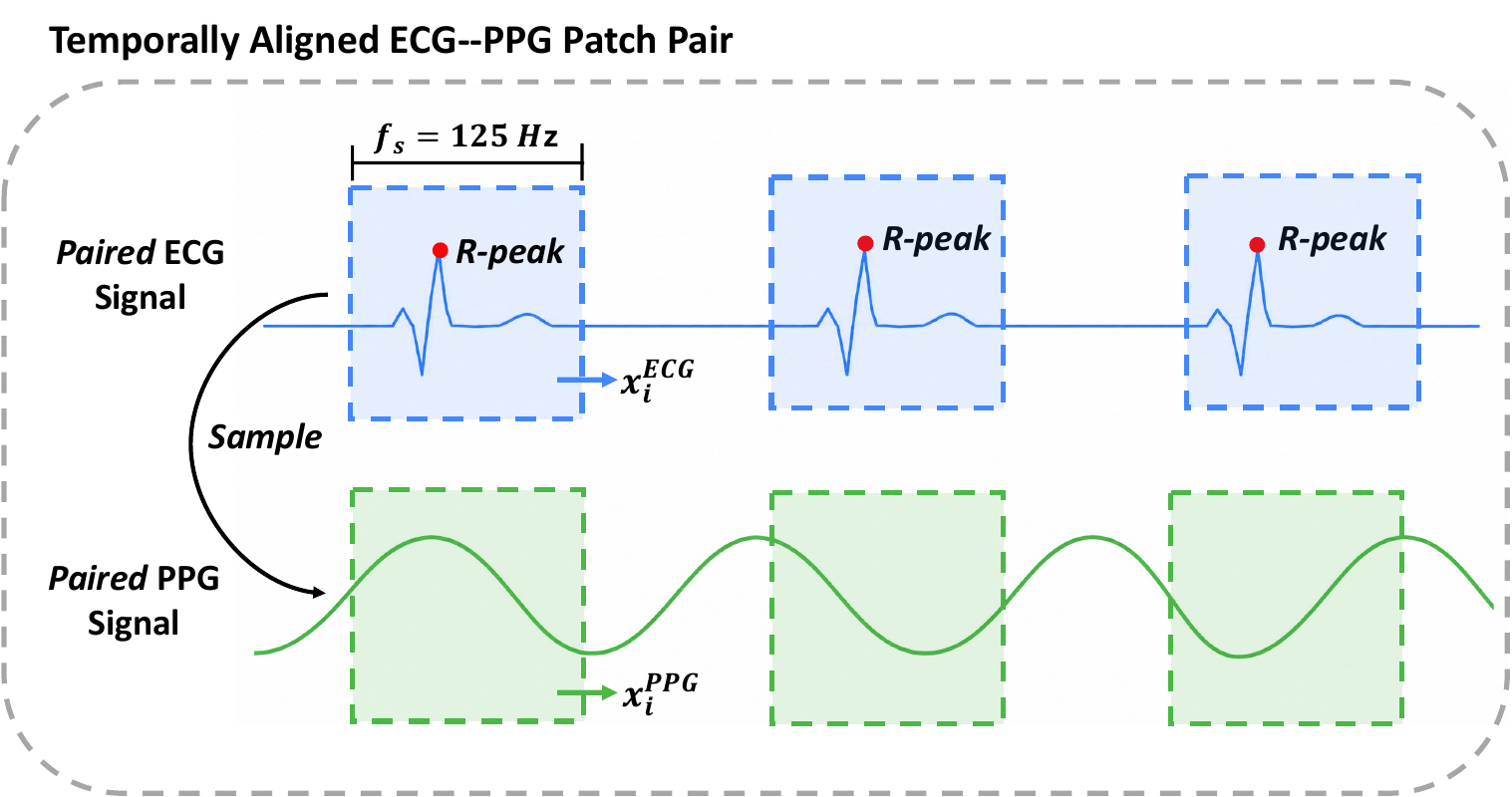}
\caption{R-peak-anchored extraction of temporally aligned ECG--PPG patch pairs. ECG R-peaks are used as temporal anchors to extract fixed-length paired ECG and PPG patches from synchronized recordings.}
\label{fig:patch_extraction}
\end{figure}

As illustrated in Fig.~\ref{fig:patch_extraction}, given 
paired ECG and PPG signals sampled at $f_s = 125\,\mathrm{Hz}$, 
R-peaks are detected from the ECG channel and used as temporal 
anchors. Around each R-peak, fixed-length windows of 
$t_{\mathrm{len}} = 125$ samples are extracted from both signals, 
producing aligned patch pairs $(x_i^{\mathrm{ECG}}, x_i^{\mathrm{PPG}})$. 
At most 20 pairs are retained per $10\,\mathrm{s}$ segment 
to control memory usage. Each PPG patch $x_i^{\mathrm{PPG}}$ is encoded by the frozen PaPaGei-S encoder~\cite{pillai_papagei_2024} and a lightweight projection head to obtain a latent PPG representation:
\begin{equation}
z_i^{\mathrm{PPG}} =
h_{\theta}\left(\phi_{\mathrm{PPG}}(x_i^{\mathrm{PPG}})\right),
\quad
z_i^{\mathrm{PPG}} \in \mathbb{R}^{d_{\mathrm{PPG}}},
\label{eq:ppg_encoder}
\end{equation}
where $\phi_{\mathrm{PPG}}(\cdot)$ denotes the frozen PaPaGei-S encoder, $h_{\theta}(\cdot)$ denotes the projection head, and $d_{\mathrm{PPG}}=64$.

We use vector quantization, implemented with CVQ-VAE~\cite{zheng_online_2023}, 
to cluster continuous PPG patch representations into discrete PPG morphology 
tokens. Each token represents a recurring local PPG pulse-shape pattern and 
is used to organize the paired ECG patches in the next phase. The PPG codebook is defined as $\mathcal{C}=\{\mathbf{c}_k\}_{k=1}^{K}$, where $K=4096$ and each code vector $\mathbf{c}*k \in \mathbb{R}^{d*{\mathrm{PPG}}}$ represents a local PPG morphology prototype.
For each 
latent PPG representation $z_i^{\mathrm{PPG}}$, vector quantization assigns it to the nearest code vector:
\begin{equation}
q(i)=
\arg\min_{k \in \{1,\ldots,K\}}
\left\|
z_i^{\mathrm{PPG}}-\mathbf{c}_k
\right\|_2 .
\label{eq:vq_assign}
\end{equation}

The selected code vector $\mathbf{c}_{q(i)}$ is then used as the quantized 
representation of the PPG patch. During training, a lightweight decoder 
$g(\cdot)$ reconstructs the original PPG patch from the selected code vector:
\begin{equation}
\hat{x}_i^{\mathrm{PPG}} = g(\mathbf{c}_{q(i)}).
\label{eq:ppg_recon}
\end{equation}

The codebook is then optimized using a reconstruction loss together with a 
commitment loss:
\begin{equation}
\mathcal{L}_{\mathrm{VQ}}
=
\left\|
x_i^{\mathrm{PPG}} - \hat{x}_i^{\mathrm{PPG}}
\right\|_2^2
+
\beta
\left\|
\mathrm{sg}[z_i^{\mathrm{PPG}}] - \mathbf{c}_{q(i)}
\right\|_2^2 ,
\label{eq:vq_loss}
\end{equation}
where $\mathrm{sg}[\cdot]$ denotes the stop-gradient operation and $\beta$ 
controls the commitment strength. This objective encourages each code vector to represent a recurring local PPG pulse-shape pattern. After training, the 
learned index $q(i)$ serves as the discrete PPG morphology token for patch 
$x_i^{\mathrm{PPG}}$.

\subsection{Phase 2: Token-Indexed ECG Retrieval Memory Construction} 

The goal of Phase~2 is to construct a token-indexed ECG retrieval 
memory that links each learned PPG morphology token to a set of 
temporally aligned ECG patch candidates from the training data.

For each paired training patch $(x_i^{\mathrm{ECG}}, x_i^{\mathrm{PPG}})$, 
the PPG patch $x_i^{\mathrm{PPG}}$ is encoded and quantized using 
the Phase~1 PPG encoding and quantization pipeline, producing its 
token assignment $q(i)$. The paired ECG patch $x_i^{\mathrm{ECG}}$ 
is then stored in the memory entry indexed by $q(i)$:
\begin{equation}
\mathcal{M}(k) = \left\{ x_i^{\mathrm{ECG}} \mid q(i)=k \right\}, 
\quad k\in\{1,\ldots,K\},
\label{eq:ecg_memory}
\end{equation}
where $\mathcal{M}(k)$ denotes the set of ECG patches associated 
with token $k$. This memory is constructed using only training 
subjects under strict subject-level splits to prevent data leakage.

In Phase~3, a query PPG patch is assigned to token $q^\ast$, and 
ECG candidates are retrieved directly from $\mathcal{M}(q^\ast)$, 
avoiding exhaustive search over the training set.

\subsection{Efficient Token-Indexed Retrieval Strategy}

Conventional retrieval-based methods identify candidate samples by nearest-neighbor search over the entire training database~\cite{cover1967nearest}, incurring a per-query complexity of $\mathcal{O}(Nd)$ for a database of $N$ patches with feature dimension $d$. Approximate schemes such as product quantization and GPU-accelerated index structures reduce this cost~\cite{jegou2010product,johnson2019billion}, but still require an explicit similarity search at query time. This becomes 
prohibitive as the training database grows.

In contrast, the proposed ETIR avoids searching over the full 
training database by using the discrete PPG token as a retrieval 
index. Given a query PPG patch $x_\ast^{\mathrm{PPG}}$, we first 
encode it into a latent representation $z_\ast^{\mathrm{PPG}}$ 
and assign it to the nearest code vector:
\begin{equation}
q^\ast = \arg\min_{k\in\{1,\ldots,K\}} 
\left\| z_\ast^{\mathrm{PPG}}-\mathbf{c}_k \right\|_2 .
\label{eq:etir_assign}
\end{equation}

The ECG candidates are then retrieved directly from the matched 
memory entry:
\begin{equation}
\mathcal{R}_\ast = \mathcal{M}(q^\ast),
\label{eq:etir_retrieve}
\end{equation}
where $\mathcal{R}_\ast$ denotes the retrieved ECG candidate set 
for the query patch. This reduces retrieval from a global search 
over $N$ training patches to a direct lookup, with a per-query 
complexity of $\mathcal{O}(Kd + |\mathcal{M}(q^\ast)|d)$. Since 
$K \ll N$ and $|\mathcal{M}(q^\ast)| \ll N$ in practice, this is 
substantially more efficient than exhaustive search.

\subsection{Phase 3: PPG-only Inference and ECG-linked Representation} 

Phase 3 performs PPG-only inference, with the goal of generating pseudo-ECG signals from test PPG segments for downstream prediction without requiring real ECG input. During PPG-only inference, detected PPG systolic peaks are used to extract 125-sample beat-level patches. In contrast, paired pretraining uses ECG R-peaks as more precise anchors because synchronized ECG is available. Although pulse transit time introduces a train--test segmentation mismatch, the codebook is designed to discretize local PPG morphology rather than to reconstruct beat-synchronous ECG.

Given the retrieved ECG candidate set $\mathcal{R}_\ast = \mathcal{M}(q^\ast)$ for a query patch, the candidates are aggregated into a single ECG patch estimate by a parameter-free average:
\begin{equation}
\hat{x}_\ast^{\mathrm{ECG}}
= \frac{1}{|\mathcal{R}_\ast|}
\sum_{x^{\mathrm{ECG}} \in \mathcal{R}_\ast} x^{\mathrm{ECG}},
\label{eq:aggregate}
\end{equation}
where up to $C$ candidates are randomly sampled without replacement when $|\mathcal{M}(q^\ast)| > C$; unless otherwise stated we use $C=100$, and we study the effect of $C$ in the ablation study (Table~\ref{tab:ablation_k}). Since this aggregation has no trainable parameters, Phase~3 introduces no additional task-specific capacity beyond the linear-probe classifier shared by all methods.

The per-patch estimates $\{\hat{x}_j^{\mathrm{ECG}}\}$ are assembled into a continuous pseudo-ECG waveform by peak-aligned overlap-add with short Hann cross-fades, using cubic-spline interpolation only to fill sparse temporal gaps. Each waveform is standardized to $1250$ samples ($10\,\mathrm{s}$ at $125\,\mathrm{Hz}$), and longer inputs are processed with $50\%$ overlapping windows.

The pseudo-ECG waveform is then upsampled from $125\,\mathrm{Hz}$ to $500\,\mathrm{Hz}$ and encoded by a frozen ECGFounder~\cite{li2024ecgfoundation}, a ResNet-based foundation model pretrained on over 10 million 12-lead ECG recordings for cardiac diagnosis, using its single-lead (lead~II) variant to obtain an ECG-linked embedding $h^{\mathrm{ECG}} \in \mathbb{R}^{512}$. In parallel, the same PPG segment is encoded by the frozen PaPaGei-S encoder~\cite{pillai_papagei_2024}, retaining its raw $512$-dimensional output, to obtain a PPG embedding $h^{\mathrm{PPG}} \in \mathbb{R}^{512}$. The final ECG-linked representation is formed by concatenation:
\begin{equation}
h^{\mathrm{final}}
= [\, h^{\mathrm{PPG}} ;\; h^{\mathrm{ECG}} \,]
\in \mathbb{R}^{1024},
\label{eq:final_repr}
\end{equation}
which is used as the input feature for downstream linear-probe classification. We adopt plain concatenation to preserve information from both modalities without introducing additional trainable parameters.

\section{Experiments}

\subsection{Datasets and Downstream Tasks}
We use the Vital subset of PulseDB~\cite{wang_pulsedb_2023} as the paired PPG--ECG training source for learning the PPG codebook and constructing the token-indexed ECG retrieval memory. This subset contains 465,480 synchronized 10-s PPG--ECG segments sampled at 125 Hz.

For downstream evaluation, we use five public physiological signal datasets covering six prediction tasks. VitalDB~\cite{lee_vitaldb_2022} is used for postoperative ICU admission prediction. Since the PulseDB Vital subset is derived from VitalDB, overlapping subjects are removed from the downstream VitalDB cohort to avoid subject-level leakage, leaving 3,753 subjects. PPG-BP~\cite{liang_new_2018} is used for hypertension classification on 205 subjects. SDB~\cite{garde_development_2014} is used for sleep-disordered breathing classification on 146 subjects. WESAD~\cite{schmidt_introducing_2018} is used for binary valence and arousal prediction using chest-worn BVP signals from 15 subjects. ECSMP~\cite{GAO2021107660} is used for mood-disturbance classification on 89 subjects. Table~\ref{tab:datasets_summary} summarizes the datasets and downstream tasks.

\begin{table}[t]
\centering
\caption{Summary of publicly available physiological signal datasets used for downstream tasks, including the target task and the number of subjects used in each cohort.}
\label{tab:datasets_summary}
\begin{tabular}{l l c}
\toprule
\textbf{Dataset} & \textbf{Task} & \textbf{No.~Subjects} \\
\midrule
VitalDB \cite{lee_vitaldb_2022} & ICU Admission & 3753 \\
PPG-BP \cite{liang_new_2018} & Hypertension & 205 \\
SDB \cite{garde_development_2014} & Sleep Disorder Breathing & 146 \\
WESAD \cite{schmidt_introducing_2018} & Valence & 15 \\
WESAD \cite{schmidt_introducing_2018} & Arousal & 15 \\
ECSMP \cite{GAO2021107660} & Mood Disturbance & 89 \\
\bottomrule
\end{tabular}
\end{table}

\subsection{Evaluation Metrics}
We use AUROC as the primary metric for all downstream classification tasks and additionally report F1 scores to assess threshold-dependent performance. For each method, 95\% confidence intervals are estimated by bootstrap resampling of the test set with 10{,}000 repetitions. For ablation studies, statistical significance is assessed using paired bootstrap tests. All reported $p$-values are two-sided and are interpreted as exploratory.

\subsection{Baseline Model}
We compare P2E-VQ with pretrained representation baselines, including REGLE~\cite{yun_unsupervised_2024}, Chronos~\cite{ansari_chronos_2024}, MOMENT~\cite{goswami_moment_2024}, and PaPaGei~\cite{pillai_papagei_2024}. For PaPaGei, both PaPaGei-P and PaPaGei-S variants are evaluated when applicable. To ensure a fair comparison, all methods are evaluated under the same frozen-feature linear-probing protocol: representations are extracted without fine-tuning, and only an $\ell_2$-regularized logistic regression classifier is trained for each downstream task. For P2E-VQ, the ECG-linked embedding is produced by a frozen ECGFounder that was pretrained on large-scale ECG data for cardiac diagnosis; this ECG encoder is used only as a fixed feature extractor and is never fine-tuned on the downstream tasks.

\subsection{Implementation Details}
All models are implemented in PyTorch and trained on a single NVIDIA RTX~4090 GPU. For codebook training, we use paired PPG--ECG patches sampled at 125~Hz with a batch size of 128 and train for 30 epochs using AdamW with a learning rate of $1\times10^{-4}$. The codebook contains $K=4096$ code vectors, each with a dimensionality of 64, and the commitment loss weight is set to $\beta=0.25$. The PPG encoder is kept frozen during codebook training.

\section{Results}
\label{sec:guidelines}

\begin{table*}[t]
\centering
\caption{Downstream comparison against pretrained models. 
95\% bootstrap CIs in brackets; best result in \textbf{bold}.}
\label{tab:auroc}
\resizebox{\textwidth}{!}{
\begin{tabular}{l|c|c|c|c|c|c}
\toprule
& \textbf{REGLE}~\cite{yun_unsupervised_2024} 
& \textbf{Chronos}~\cite{ansari_chronos_2024} 
& \textbf{Moment}~\cite{goswami_moment_2024} 
& \textbf{PaPaGei-P}~\cite{pillai_papagei_2024} 
& \textbf{PaPaGei-S}~\cite{pillai_papagei_2024} 
& \textbf{P2E-VQ (Ours)} \\
\midrule
\multicolumn{7}{l}{\textbf{AUROC} ($\uparrow$)} \\
\hline
ICU Admission 
& 0.57 \smat{[0.52-0.62]} 
& 0.73 \smat{[0.68-0.80]} 
& 0.72 \smat{[0.70-0.80]} 
& 0.73 \smat{[0.67-0.78]} 
& 0.73 \smat{[0.70-0.76]} 
& \textbf{0.75} \smat{[0.72-0.78]} \\
Hypertension 
& 0.47 \smat{[0.34-0.58]} 
& 0.57 \smat{[0.43-0.71]} 
& 0.75 \smat{[0.64-0.85]} 
& 0.74 \smat{[0.55-0.90]} 
& 0.77 \smat{[0.68-0.87]} 
& \textbf{0.81} \smat{[0.71-0.91]} \\
Sleep Disordered Breathing 
& 0.45 \smat{[0.30-0.61]} 
& 0.58 \smat{[0.35-0.82]} 
& 0.45 \smat{[0.23-0.66]} 
& 0.54 \smat{[0.23-0.66]} 
& 0.56 \smat{[0.39-0.71]} 
& \textbf{0.62} \smat{[0.44-0.80]} \\
Mood Disturbance 
& 0.41 \smat{[0.16-0.66]} 
& 0.43 \smat{[0.21-0.68]} 
& 0.55 \smat{[0.33-0.78]} 
& 0.53 \smat{[0.27-0.78]} 
& 0.59 \smat{[0.34-0.82]} 
& \textbf{0.63} \smat{[0.40-0.85]} \\
Valence 
& 0.55 \smat{[0.52-0.57]} 
& 0.56 \smat{[0.53-0.59]} 
& 0.57 \smat{[0.54-0.59]} 
& 0.53 \smat{[0.51-0.56]} 
& 0.66 \smat{[0.63-0.69]} 
& \textbf{0.68} \smat{[0.65-0.71]} \\
Arousal 
& 0.55 \smat{[0.52-0.58]} 
& 0.57 \smat{[0.54-0.60]} 
& 0.56 \smat{[0.53-0.58]} 
& 0.58 \smat{[0.55-0.61]} 
& 0.68 \smat{[0.65-0.71]} 
& \textbf{0.72} \smat{[0.69-0.75]} \\
\hline
\rowcolor[gray]{.90}
Average 
& 0.50 $\pm$ 0.06 
& 0.57 $\pm$ 0.09 
& 0.60 $\pm$ 0.10 
& 0.61 $\pm$ 0.09 
& 0.67 $\pm$ 0.07 
& \textbf{0.70 $\pm$ 0.07} \\
\midrule
\multicolumn{7}{l}{\textbf{F1} ($\uparrow$)} \\
\hline
ICU Admission 
& 0.00 \smat{[0.00-0.00]} 
& 0.20 \smat{[0.11-0.30]} 
& 0.12 \smat{[0.04-0.20]} 
& 0.12 \smat{[0.04-0.20]} 
& \textbf{0.26} \smat{[0.18-0.33]} 
& 0.19 \smat{[0.14-0.25]} \\
Hypertension 
& 0.77 \smat{[0.70-0.84]} 
& 0.68 \smat{[0.58-0.77]} 
& 0.75 \smat{[0.66-0.84]} 
& \textbf{0.84} \smat{[0.72-0.92]} 
& 0.78 \smat{[0.70-0.86]} 
& 0.82 \smat{[0.75-0.89]} \\
Sleep Disordered Breathing 
& 0.00 \smat{[0.00-0.00]} 
& 0.33 \smat{[0.00-0.60]} 
& 0.22 \smat{[0.00-0.47]} 
& 0.32 \smat{[0.00-0.60]} 
& 0.47 \smat{[0.23-0.67]} 
& \textbf{0.51} \smat{[0.31-0.68]} \\
Mood Disturbance 
& 0.00 \smat{[0.00-0.00]} 
& 0.36 \smat{[0.10-0.59]} 
& 0.23 \smat{[0.00-0.47]} 
& 0.37 \smat{[0.00-0.66]} 
& 0.44 \smat{[0.13-0.69]} 
& \textbf{0.51} \smat{[0.25-0.73]} \\
Valence 
& 0.00 \smat{[0.00-0.00]} 
& 0.10 \smat{[0.07-0.14]} 
& 0.12 \smat{[0.09-0.16]} 
& 0.17 \smat{[0.13-0.21]} 
& 0.09 \smat{[0.06-0.13]} 
& \textbf{0.34} \smat{[0.29-0.38]} \\
Arousal 
& 0.83 \smat{[0.81-0.84]} 
& 0.82 \smat{[0.80-0.83]} 
& 0.81 \smat{[0.79-0.82]} 
& 0.81 \smat{[0.79-0.82]} 
& 0.88 \smat{[0.87-0.89]} 
& 0.87 \smat{[0.86-0.89]} \\
\hline
\rowcolor[gray]{.90}
Average 
& 0.27 $\pm$ 0.38 
& 0.42 $\pm$ 0.25 
& 0.38 $\pm$ 0.29 
& 0.44 $\pm$ 0.29 
& 0.49 $\pm$ 0.27 
& \textbf{0.54 $\pm$ 0.24} \\
\bottomrule
\end{tabular}}
\end{table*}

\subsection{Downstream Task Performance}

Table~\ref{tab:auroc} reports the downstream performance of
P2E-VQ and pretrained representation baselines under the same
frozen-feature linear-probing protocol. AUROC is used as the primary
metric, and F1 score is reported as a threshold-dependent measure.
Across the six downstream tasks, P2E-VQ achieves the highest
average AUROC (0.70 $\pm$ 0.07) and the highest average F1 score
(0.54 $\pm$ 0.24). It also obtains the best AUROC on all evaluated
endpoints.

Compared with PaPaGei-S, the strongest baseline on average, P2E-VQ
improves AUROC from 0.73 to 0.75 for ICU admission, from 0.77 to
0.81 for hypertension, from 0.56 to 0.62 for sleep-disordered
breathing, from 0.59 to 0.63 for mood disturbance, from 0.66 to
0.68 for valence, and from 0.68 to 0.72 for arousal. The consistent
AUROC gains across all endpoints indicate that ECG-linked retrieval
improves the ranking quality of PPG representations beyond the
PPG-only pretrained baseline.

The improvements are particularly evident for sleep-disordered
breathing and mood disturbance, where PaPaGei-S shows relatively
lower AUROC performance. On these two tasks, P2E-VQ improves AUROC
by 0.06 and 0.04, respectively, while also increasing F1 score from
0.47 to 0.51 and from 0.44 to 0.51. For arousal prediction, P2E-VQ
improves AUROC from 0.68 to 0.72 while maintaining a comparable F1
score to PaPaGei-S (0.87 vs.\ 0.88). These results indicate that
ECG-linked retrieval provides additional discriminative information
when PPG-only representations are less separable.

For the clinically grounded endpoints, P2E-VQ improves both AUROC
and F1 score on hypertension, from 0.77 to 0.81 and from 0.78 to
0.82, respectively. This result indicates that the ECG-linked
representation provides complementary information for cardiovascular
risk-related prediction. On ICU admission, P2E-VQ achieves the
highest AUROC, but its F1 score is lower than that of PaPaGei-S
(0.19 vs.\ 0.26), indicating that the gain in ranking performance does
not necessarily translate into improved threshold-dependent
classification under class imbalance.

For the affective endpoints, P2E-VQ also improves AUROC over
PaPaGei-S. In valence prediction, AUROC increases from 0.66 to 0.68,
while F1 score increases from 0.09 to 0.34. In arousal prediction,
AUROC increases from 0.68 to 0.72, with a comparable F1 score
(0.87 vs.\ 0.88). These results indicate that the benefit of
ECG-linked retrieval is not limited to clinical endpoints, but also
extends to affect-related prediction tasks.

Overall, these results support the main conclusion of this work:
P2E-VQ improves PPG-only downstream prediction by augmenting PPG
representations with ECG-linked retrieval rather than relying on
deterministic ECG waveform reconstruction. The wider confidence
intervals observed on WESAD and ECSMP are consistent with their
smaller cohort sizes.

\subsection{Retrieval Fidelity of ECG-linked Candidates}

Since P2E-VQ retrieves ECG candidates according to PPG token
assignments, we evaluate whether token-matched retrieval produces ECG
patches more similar to the true paired ECG than retrieval from an
unmatched memory entry. For each held-out paired PPG--ECG patch, the
PPG patch is assigned to a codebook token, and ECG candidates are
retrieved from the corresponding memory entry. As a shuffled control,
the matched entry is replaced with a randomly selected different
codebook entry, and ECG candidates are aggregated in the same way.

As shown in Table~\ref{tab:fidelity}, matched retrieval consistently
outperforms shuffled retrieval across all waveform-similarity metrics:
cosine similarity (0.889 vs.\ 0.783, Cohen's $d = 0.90$), Pearson
correlation (0.739 vs.\ 0.479, Cohen's $d = 0.79$), and $\ell_2$
distance (1.508 vs.\ 1.987). All differences are statistically
significant ($p < 10^{-300}$), and matched retrieval is superior in
86.2\% of held-out patches by cosine similarity.

These results indicate that the learned PPG codebook organizes PPG
patches in a way that preserves ECG-linked structure. Consequently,
token-matched retrieval produces ECG candidates that are structurally
closer to the true paired ECG than unmatched code-level retrieval,
supporting their use as complementary information for downstream
PPG-only prediction.

\begin{figure*}[t]
\centering
\includegraphics[width=0.9\linewidth]{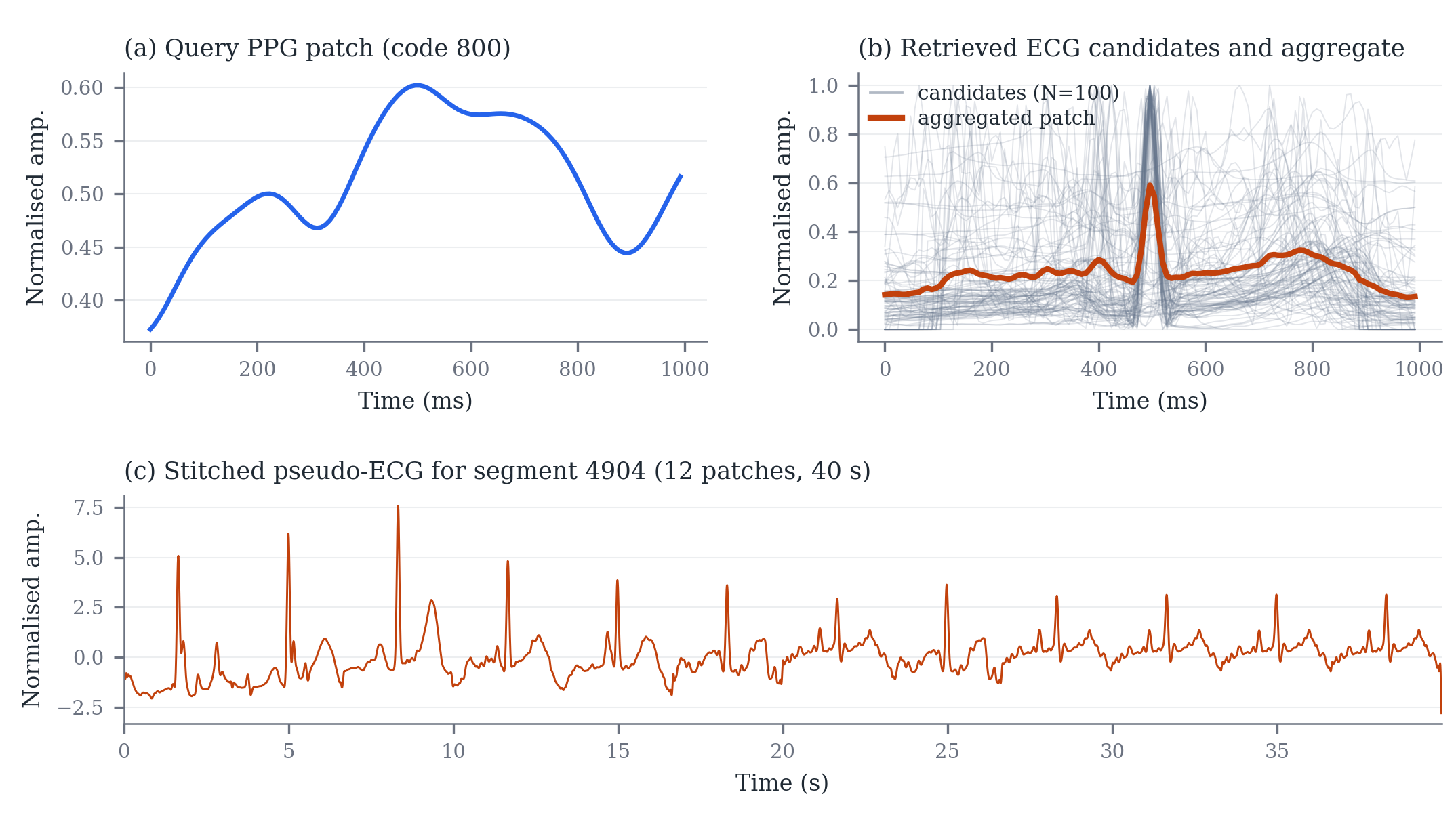}
\caption{Qualitative example of token-indexed retrieval on a WESAD test segment under PPG-only inference.
(a) A query PPG patch extracted at a systolic peak and quantized to codebook entry 800.
(b) The $N=100$ real ECG patches stored under that entry (grey) together with their parameter-free aggregate (orange, Eq.~\eqref{eq:aggregate}). The candidates originate from different training subjects, yet their QRS complexes fall at consistent positions within the patch window, so averaging retains a distinct R-peak and T-wave instead of cancelling them.
(c) The per-patch aggregates assembled by peak-aligned overlap-add into a 40-s pseudo-ECG for the same segment, which is subsequently split into overlapping 10-s windows for encoding. The waveform serves only as an input to the frozen ECG encoder and is not a subject-specific ECG reconstruction.}
\label{fig:retrieval_case}
\end{figure*}

Fig.~\ref{fig:retrieval_case} illustrates this behaviour on a representative test segment. The query PPG patch in Fig.~\ref{fig:retrieval_case}(a) is quantized to a single codebook entry whose memory holds 100 real ECG patches contributed by different training subjects. Although these candidates differ in amplitude and baseline, their QRS complexes remain temporally consistent within the patch window, so the parameter-free average in Fig.~\ref{fig:retrieval_case}(b) preserves a distinct R-peak and a visible T-wave rather than collapsing into a flat template. Assembling the per-patch aggregates by overlap-add produces the pseudo-ECG in Fig.~\ref{fig:retrieval_case}(c), whose beat structure follows the pulse rhythm of the input PPG. This is the qualitative counterpart of the fidelity gap in Table~\ref{tab:fidelity}: retrieval keyed on PPG tokens returns ECG morphology that is structurally coherent, which is what makes the aggregated patch usable as input to the frozen ECG encoder.

\begin{table}[]
\centering
\caption{Retrieval fidelity on held-out paired PPG--ECG patches.}
\label{tab:fidelity}
\footnotesize
\setlength{\tabcolsep}{3.5pt}
\begin{tabular}{lccc}
\hline
\textbf{Metric} & \textbf{Retrieved} & \textbf{Shuffled} & \textbf{$\Delta$} \\
\hline
Cosine sim. ($\uparrow$)
& $0.889 \pm 0.097$
& $0.783 \pm 0.140$
& $+0.106$ \\
Pearson $r$ ($\uparrow$)
& $0.739 \pm 0.283$
& $0.479 \pm 0.368$
& $+0.260$ \\
$\ell_2$ dist. ($\downarrow$)
& $1.508 \pm 0.827$
& $1.987 \pm 0.884$
& $-0.479$ \\
\hline
\multicolumn{4}{l}{$n=561{,}910$ held-out patches.} \\
\multicolumn{4}{l}{Wilcoxon signed-rank test: $p < 10^{-300}$ for all metrics.} \\
\multicolumn{4}{l}{Cohen's $d$: 0.90 for cosine similarity and 0.79 for Pearson $r$.} \\
\multicolumn{4}{l}{Retrieved $>$ shuffled in 86.2\% of patches by cosine similarity.} \\
\hline
\end{tabular}
\end{table}

\section{Ablation study}
\subsection{Effect of ECG-linked Content}
We conduct these controlled analyses on WESAD because it shows the largest ECG-linked improvement among all tasks (Table~\ref{tab:auroc}), making it the most sensitive setting for isolating the contribution of ECG-linked content; extending these controls to larger cohorts is noted as a limitation. Because the proposed fusion concatenates the 512-dimensional PPG embedding with an additional ECG-linked embedding, the performance gain could be confounded by the increased feature dimensionality. We therefore conduct a matched-capacity control study to determine whether the gains arise from ECG-linked information rather than dimensionality or classifier-capacity effects. Specifically, we compare the aligned ECG-linked fusion with a stronger PPG-only baseline and two controls. The stronger baseline is a 1024-dimensional PPG representation obtained by concatenating two distinct PaPaGei-S embeddings, namely the projected features used throughout this work and the pooled features preceding the projection head. The two controls are a shuffled-test setting that breaks the sample-wise PPG--ECG correspondence at inference, and a PPG-duplicate setting that matches the fused feature dimensionality by concatenating the PPG embedding with itself.
\begin{table*}[]
\centering
\caption{Control experiments separating ECG-linked content from feature dimensionality in valence and arousal prediction.}
\label{tab:wesad_controls}
\setlength{\tabcolsep}{6pt}
\begin{tabular}{lcccccc}
\toprule
& \multicolumn{3}{c}{\textbf{Valence}} 
& \multicolumn{3}{c}{\textbf{Arousal}} \\
\cmidrule(lr){2-4} \cmidrule(lr){5-7}
Condition & AUROC & $\Delta$ & $p$ 
          & AUROC & $\Delta$ & $p$ \\
\midrule
PPG only (512-d) 
  & 0.664 & -- & -- 
  & 0.677 & -- & -- \\
PPG only (1024-d) 
  & 0.673 & $+$0.009 & -- 
  & 0.687 & $+$0.010 & -- \\
PPG + pseudo-ECG (aligned, 1024-d) 
  & \textbf{0.679} & $+$0.015 & 0.12 
  & \textbf{0.723} & $+$0.046 & 0.002 \\
PPG + pseudo-ECG (shuffled at test) 
  & 0.604 & $-$0.060 & ${<}0.001$ 
  & 0.633 & $-$0.044 & ${<}0.001$ \\
PPG + PPG duplicate (1024-d) 
  & 0.659 & $-$0.005 & -- 
  & 0.681 & $+$0.004 & -- \\
\bottomrule
\end{tabular}
\end{table*}

As shown in Table~\ref{tab:wesad_controls}, aligned ECG-linked fusion improves AUROC over the PPG-only baseline from 0.664 to 0.679 for valence and from 0.677 to 0.723 for arousal. In contrast, the shuffled-test setting decreases AUROC to 0.604 and 0.633, respectively, showing that breaking the sample-wise PPG--ECG correspondence degrades performance. The PPG-duplicate setting, despite having the same 1024-dimensional input as the aligned fusion, shows no meaningful gain over PPG only. The genuinely 1024-dimensional PPG baseline, whose additional half is a near-orthogonal PaPaGei-S embedding rather than a copy, improves arousal AUROC by only 0.010, compared with 0.047 for ECG-linked fusion. These results indicate that the improvement is driven by ECG-linked content rather than feature dimensionality, feature duplication, or additional PPG representational capacity.

A natural objection is that the retrieved patches belong to other subjects and are aggregated into a population-level prototype, which may appear to reproduce the very criticism we raise against waveform reconstruction. We therefore make the distinction explicit. Our objection to parametric reconstruction is not that it fails to recover an individual's true ECG, which is unattainable from PPG alone, but that it can synthesize morphologies that were never observed in any recording. Retrieval cannot hallucinate in this sense: every item returned by the memory is a real ECG segment measured in a training subject. Nor is the retrieved content an undifferentiated population average. Were it so, permuting the retrieved ECG across test samples would leave performance unchanged; instead it lowers AUROC to 0.604 and 0.633, below the PPG-only baseline itself (Table~\ref{tab:wesad_controls}), and retrieved patches are closer to the true paired ECG than shuffled ones in 86.2\% of cases (Table~\ref{tab:fidelity}). The memory thus supplies token-conditioned ECG evidence rather than a single global prototype. We consequently frame P2E-VQ as representation-level transfer, and we do not claim, and do not require, subject-specific ECG fidelity.

\subsection{Effect of the Number of Retrieved Candidates}
To evaluate the sensitivity of P2E-VQ to the retrieval pool size, we vary the number of retrieved ECG candidates $C$ and compare downstream performance.

\begin{table}[]
\centering
\caption{Effect of the number of retrieved ECG candidates
         on valence and arousal prediction.}
\label{tab:ablation_k}
\footnotesize
\setlength{\tabcolsep}{4pt}
\begin{tabular}{@{}lcccc@{}}
\toprule
\textbf{$C$} 
  & \multicolumn{2}{c}{\textbf{Valence}} 
  & \multicolumn{2}{c}{\textbf{Arousal}} \\
\cmidrule(lr){2-3} \cmidrule(lr){4-5}
& AUROC & $\Delta$ & AUROC & $\Delta$ \\
\midrule
0 (PPG only) & 0.664 & --           & 0.677 & --           \\
20           & 0.662 & $-$0.002     & 0.720 & $+$0.043     \\
50           & 0.676 & $+$0.012     & 0.723 & $+$0.046     \\
200          & 0.679 & $+$0.015     & \textbf{0.727} & \textbf{$+$0.050} \\
500          & \textbf{0.680} & \textbf{$+$0.016} 
             & \textbf{0.727} & \textbf{$+$0.050} \\
\bottomrule
\end{tabular}
\end{table}

As shown in Table~\ref{tab:ablation_k}, arousal prediction improves over the PPG-only baseline for all tested values of $C$, while valence prediction improves once $C \geq 50$. Increasing $C$ from 50 to 200 or 500 yields only marginal additional gains, indicating that most of the benefit is obtained with a moderate candidate pool. These results suggest that P2E-VQ is not highly sensitive to the exact choice of $C$ once sufficient ECG candidates are available for aggregation.

\section{Efficiency of Token-indexed Retrieval}
To assess the computational benefit of token-indexed retrieval, we
compare the retrieval cost of P2E-VQ with a brute-force k-nearest-neighbour (kNN)\cite{cover1967nearest} baseline
that searches over all training PPG embeddings.

In P2E-VQ, each query patch is assigned to one of $K{=}4096$ codebook
entries, and ECG candidates are retrieved directly from the matched
memory entry. In contrast, brute-force kNN requires comparing the query
against approximately $5.6{\times}10^6$ training embeddings. This
reduces the number of distance comparisons by approximately
$1367\times$. The retrieval index is also much smaller: the VQ codebook
requires about 1 MB, whereas storing all training embeddings for kNN
requires about 1.4 GB. In empirical timing on a single NVIDIA
RTX~4090 GPU, token-indexed retrieval processes a 10 s PPG segment in
approximately 15 ms, compared with approximately 200 ms for brute-force
kNN.

These results show that token-indexed retrieval substantially reduces
retrieval cost while preserving the non-parametric ECG-linked retrieval
mechanism.

\section{Limitation}
P2E-VQ has several limitations. First, the retrieved ECG candidates are aggregated across training subjects that share the same PPG token, so the resulting pseudo-ECG represents a population-level ECG prior rather than a subject-specific reconstruction of an individual's true ECG. It is therefore intended as a representation-level cue for downstream prediction and is not suitable for direct ECG-based clinical diagnosis.

Second, paired pretraining anchors patches at ECG R-peaks, whereas PPG-only inference anchors at systolic peaks. The resulting pulse-transit-time offset, together with the reliance on PPG peak detection, means that retrieval precision may degrade under heavy motion artifacts, low signal-to-noise conditions, or irregular rhythms such as ectopic beats and atrial fibrillation, where peak detection is unreliable. A signal-quality index or fallback segmentation could improve robustness in these settings.

Third, the evaluation has several scope constraints. On the largest cohort (ICU admission), P2E-VQ improves AUROC but not F1, indicating that gains in ranking do not necessarily translate into threshold-dependent performance under class imbalance; precision--recall analyses such as AUPRC would provide a fuller picture. The matched-capacity controls and the retrieval-fidelity analysis are reported only on WESAD, and the fidelity is measured on ECG-anchored held-out patches, so it characterizes how the codebook organizes ECG-linked structure rather than the retrieval quality under PPG-only inference. All $p$-values are exploratory and uncorrected, and the smaller cohorts use a single held-out split rather than leave-one-subject-out cross-validation.

Finally, codebook and memory construction require synchronized PPG--ECG recordings, which restricts applicability to settings where paired data are available. All evaluation datasets use clinical- or research-grade sensors, including chest-worn BVP and fingertip PPG; robustness across consumer wrist-worn PPG, diverse skin tones, and ambulatory conditions remains untested, and the downstream tasks are binary. Because the memory stores real ECG patches from training subjects, privacy safeguards should also be considered before clinical deployment.

We also note that this work does not include a head-to-head comparison against parametric PPG-to-ECG reconstruction models such as CardioGAN or RDDM under our downstream protocol. Our claim is therefore that retrieval provides a viable alternative that avoids the ill-posed inverse mapping, not that it is empirically superior to reconstruction for every task; establishing the latter requires a direct comparison and is left for future work.

\section{Conclusion}
We presented P2E-VQ, a retrieval-augmented framework that transfers ECG-linked information into PPG representations through discrete patch tokenization and a token-indexed ECG memory, enabling ECG-linked inference from PPG-only recordings without subject-specific ECG reconstruction. Its efficient token-indexed retrieval (ETIR) strategy fetches ECG candidates directly from the matched memory entry, reducing the number of distance comparisons by roughly $1367\times$ relative to brute-force nearest-neighbour search.

Under a unified frozen-feature linear-probing protocol across five datasets and six downstream tasks, P2E-VQ yields consistent improvements over strong pretrained baselines (average AUROC 0.70 versus 0.67 for the strongest baseline), and matched-capacity negative controls confirm that the gains stem from ECG-linked content rather than increased feature dimensionality. The benefit is task-dependent, being most pronounced for endpoints where ECG morphology provides information complementary to PPG. Future work includes validation on consumer wrist-worn PPG, ablation of the codebook size and ECG encoder, and signal-quality gating for robustness in arrhythmia-prone populations.

\section{References}

\bibliographystyle{IEEEtran}
\bibliography{reference}

@article{li2024ecgfoundation,
  author       = {Jun Li and
                  Aaron Aguirre and
                  Junior Moura and
                  Che Liu and
                  Lanhai Zhong and
                  Chenxi Sun and
                  Gari D. Clifford and
                  M. Brandon Westover and
                  Shenda Hong},
  title        = {An Electrocardiogram Foundation Model Built on over 10 Million Recordings
                  with External Evaluation across Multiple Domains},
  journal      = {CoRR},
  volume       = {abs/2410.04133},
  year         = {2024},
  url          = {https://doi.org/10.48550/arXiv.2410.04133},
  doi          = {10.48550/ARXIV.2410.04133},
  eprinttype   = {arXiv},
  eprint       = {2410.04133},
  bibsource    = {dblp computer science bibliography, https://dblp.org}
}

@inproceedings{vo2021p2ewgan,
author = {Vo, Khuong and Naeini, Emad Kasaeyan and Naderi, Amir and Jilani, Daniel and Rahmani, Amir M. and Dutt, Nikil and Cao, Hung},
title = {P2E-WGAN: ECG waveform synthesis from PPG with conditional wasserstein generative adversarial networks},
year = {2021},
isbn = {9781450381048},
}

@article{GAO2021107660,
title = {ECSMP: A dataset on emotion, cognition, sleep, and multi-model physiological signals},
journal = {Data in Brief},
volume = {39},
pages = {107660},
year = {2021},
issn = {2352-3409},
doi = {https://doi.org/10.1016/j.dib.2021.107660},
url = {https://www.sciencedirect.com/science/article/pii/S2352340921009355},
author = {Zhilin Gao and Xingran Cui and Wang Wan and Wenming Zheng and Zhongze Gu}
}

@inproceedings{banerjee2014photoecg,
  author    = {Banerjee, Riju and Sinha, Atanu and Choudhury, A. D. and Visvanathan, A.},
  title     = {{PhotoECG}: Photoplethysmography to Estimate {ECG} Parameters},
  booktitle = {Proc. IEEE Int. Conf. Acoustics, Speech and Signal Processing (ICASSP)},
  year      = {2014}
}

@inproceedings{tian2020crossdomain,
  author    = {Tian, Xin and Zhu, Qiang and Li, Yuhui and Wu, Min},
  title     = {Cross-Domain Joint Dictionary Learning for {ECG} Reconstruction from {PPG}},
  booktitle = {Proc. IEEE Int. Conf. Acoustics, Speech and Signal Processing (ICASSP)},
  year      = {2020}
}

@article{zhang2025ecgfm,
  title={ECGFM: A foundation model for ECG analysis trained on a multi-center million-ECG dataset},
  author={Zhang, Shaoting and Du, Yishan and Wang, Wenji and He, Xianying and Cui, Fangfang and Zhao, Liang and Wang, Bei and Hu, Zhiqiang and Wang, Ziqiang and Xia, Qing and others},
  journal={Information Fusion},
  pages={103363},
  year={2025},
  publisher={Elsevier}
}

@misc{khandelwal2021nearestneighbormachinetranslation,
      title={Nearest Neighbor Machine Translation}, 
      author={Urvashi Khandelwal and Angela Fan and Dan Jurafsky and Luke Zettlemoyer and Mike Lewis},
      year={2021},
      eprint={2010.00710},
      archivePrefix={arXiv},
      primaryClass={cs.CL},
      url={https://arxiv.org/abs/2010.00710}, 
}

@article{jegou2010product,
  title={Product quantization for nearest neighbor search},
  author={Jegou, Herve and Douze, Matthijs and Schmid, Cordelia},
  journal={IEEE transactions on pattern analysis and machine intelligence},
  volume={33},
  number={1},
  pages={117--128},
  year={2010},
  publisher={IEEE}
}

@article{ho2022quickly,
  author  = {Ho, Wen-Hsien and others},
  title   = {Quickly Convert Photoplethysmography to Electrocardiogram Signals},
  journal = {IEEE Access},
  year    = {2022}
}

@article{tang2022robust,
  author  = {Tang, Qiang and Chen, Zhou and Guo, Yujun and Liang, Yongbo and Ward, Rabab and Menon, Carlo and Elgendi, Mohamed},
  title   = {Robust Reconstruction of Electrocardiogram Using Photoplethysmography: A Subject-Based Model},
  journal = {Frontiers in Physiology},
  year    = {2022},
  volume  = {13}
}

@article{tang2023ppg2ecgps,
  title={PPG2ECGps: An End-to-End Subject-Specific Deep Neural Network Model for Electrocardiogram Reconstruction from Photoplethysmography Signals without Pulse Arrival Time Adjustments},
  author={Qunfeng Tang and Zhencheng Chen and Rabab Kreidieh Ward and Carlo Menon and Mohamed Elgendi},
  journal={Bioengineering},
  year={2023},
  volume={10},
  url={https://api.semanticscholar.org/CorpusID:258878351}
}

@article{ezzat2024hybrid,
  author  = {Ezzat, Sameh and Abdel-Raheem, Effat and Abd El-Samie, Fathi E. and others},
  title   = {{ECG} Signal Reconstruction from {PPG} Using a Hybrid Attention-Based Deep Learning Network},
  journal = {EURASIP Journal on Advances in Signal Processing},
  year    = {2024}
}

@inproceedings{sarkar2021cardiogan,
  author    = {Sarkar, Pritam and Etemad, Ali},
  title     = {{CardioGAN}: Attentive Generative Adversarial Network with Dual Discriminators for Synthesis of {ECG} from {PPG}},
  booktitle = {Proc. AAAI Conf. Artificial Intelligence},
  year      = {2021}
}

@article{li2024biodiffusion,
  author  = {Li, Xiaomin and Sakevych, Mykhailo and Atkinson, Gentry and Metsis, Vangelis},
  title   = {{BioDiffusion}: A Versatile Diffusion Model for Biomedical Signal Synthesis},
  journal = {Bioengineering},
  year    = {2024},
  volume  = {11},
  number  = {4},
  articleno = {299}
}

@inproceedings{oord2017neural,
  author    = {van den Oord, Aaron and Vinyals, Oriol and Kavukcuoglu, Koray},
  title     = {Neural Discrete Representation Learning},
  booktitle = {Advances in Neural Information Processing Systems (NeurIPS)},
  year      = {2017}
}

@article{sun2005beatbased,
  author  = {Sun, Chih-Chin and Tai, Shen-Chuan},
  title   = {Beat-Based {ECG} Compression Using Gain-Shape Vector Quantization},
  journal = {IEEE Transactions on Biomedical Engineering},
  volume  = {52},
  number  = {11},
  pages   = {1882--1888},
  year    = {2005},
  doi     = {10.1109/TBME.2005.856270}
}

@article{sarkar2022selfsupervised,
   title={Self-Supervised ECG Representation Learning for Emotion Recognition},
   volume={13},
   ISSN={2371-9850},
   url={http://dx.doi.org/10.1109/TAFFC.2020.3014842},
   DOI={10.1109/taffc.2020.3014842},
   number={3},
   journal={IEEE Transactions on Affective Computing},
   publisher={Institute of Electrical and Electronics Engineers (IEEE)},
   author={Sarkar, Pritam and Etemad, Ali},
   year={2022},
   month=July, pages={1541–1554} }

@inproceedings{lewis2020rag,
  author    = {Lewis, Patrick and Perez, Ethan and Piktus, Aleksandra and Petroni, Fabio and Karpukhin, Vladimir and Goyal, Naman and others},
  title     = {Retrieval-Augmented Generation for Knowledge-Intensive {NLP} Tasks},
  booktitle = {Advances in Neural Information Processing Systems (NeurIPS)},
  year      = {2020}
}

@inproceedings{zhu2019ecg,
  title={ECG reconstruction via PPG: A pilot study},
  author={Zhu, Qiang and Tian, Xin and Wong, Chau-Wai and Wu, Min},
  booktitle={2019 IEEE EMBS international conference on biomedical \& health informatics (BHI)},
  pages={1--4},
  year={2019},
  organization={IEEE}
}

@article{gil_photoplethysmography_2010,
	title = {Photoplethysmography pulse rate variability as a surrogate measurement of heart rate variability during non-stationary conditions},
	volume = {31},
	issn = {1361-6579},
	doi = {10.1088/0967-3334/31/9/015},
	language = {eng},
	number = {9},
	journal = {Physiological Measurement},
	author = {Gil, E. and Orini, M. and Bailón, R. and Vergara, J. M. and Mainardi, L. and Laguna, P.},
	month = sep,
	year = {2010},
	pmid = {20702919},
	pages = {1271--1290},
}

@article{schafer_how_2013,
	title = {How accurate is pulse rate variability as an estimate of heart rate variability? {A} review on studies comparing photoplethysmographic technology with an electrocardiogram},
	volume = {166},
	issn = {1874-1754},
	shorttitle = {How accurate is pulse rate variability as an estimate of heart rate variability?},
	doi = {10.1016/j.ijcard.2012.03.119},
	language = {eng},
	number = {1},
	journal = {International Journal of Cardiology},
	author = {Schäfer, Axel and Vagedes, Jan},
	month = jun,
	year = {2013},
	pmid = {22809539},
	pages = {15--29},
}

@article{kligfield_ecg_standard_2007,
author = {Paul Kligfield  and Leonard S. Gettes  and James J. Bailey  and Rory Childers  and Barbara J. Deal  and E. William Hancock  and Gerard van Herpen  and Jan A. Kors  and Peter Macfarlane  and David M. Mirvis  and Olle Pahlm  and Pentti Rautaharju  and Galen S. Wagner },
title = {Recommendations for the Standardization and Interpretation of the Electrocardiogram},
journal = {Circulation},
volume = {115},
number = {10},
pages = {1306-1324},
year = {2007},
doi = {10.1161/CIRCULATIONAHA.106.180200},
URL = {https://www.ahajournals.org/doi/abs/10.1161/CIRCULATIONAHA.106.180200},
eprint = {https://www.ahajournals.org/doi/pdf/10.1161/CIRCULATIONAHA.106.180200}}

@misc{zheng_online_2023,
	title = {Online {Clustered} {Codebook}},
	url = {http://arxiv.org/abs/2307.15139},
	doi = {10.48550/arXiv.2307.15139},
	urldate = {2025-10-31},
	publisher = {arXiv},
	author = {Zheng, Chuanxia and Vedaldi, Andrea},
	month = jul,
	year = {2023},
	note = {arXiv:2307.15139 [cs]},
}

@article{liang_new_2018,
	title = {A new, short-recorded photoplethysmogram dataset for blood pressure monitoring in {China}},
	volume = {5},
	copyright = {2018 The Author(s)},
	issn = {2052-4463},
	url = {https://www.nature.com/articles/sdata201820},
	doi = {10.1038/sdata.2018.20},
	language = {en},
	number = {1},
	urldate = {2025-10-23},
	journal = {Scientific Data},
	author = {Liang, Yongbo and Chen, Zhencheng and Liu, Guiyong and Elgendi, Mohamed},
	month = feb,
	year = {2018},
	note = {Publisher: Nature Publishing Group},
	pages = {180020},
}

@inproceedings{schmidt_introducing_2018,
	address = {New York, NY, USA},
	series = {{ICMI} '18},
	title = {Introducing {WESAD}, a {Multimodal} {Dataset} for {Wearable} {Stress} and {Affect} {Detection}},
	isbn = {978-1-4503-5692-3},
	url = {https://doi.org/10.1145/3242969.3242985},
	doi = {10.1145/3242969.3242985},
	urldate = {2025-10-22},
	booktitle = {Proceedings of the 20th {ACM} {International} {Conference} on {Multimodal} {Interaction}},
	publisher = {Association for Computing Machinery},
	author = {Schmidt, Philip and Reiss, Attila and Duerichen, Robert and Marberger, Claus and Van Laerhoven, Kristof},
	year = {2018},
	pages = {400--408},
}

@article{wang_pulsedb_2023,
	title = {{PulseDB}: {A} large, cleaned dataset based on {MIMIC}-{III} and {VitalDB} for benchmarking cuff-less blood pressure estimation methods},
	volume = {4},
	issn = {2673-253X},
	shorttitle = {{PulseDB}},
	url = {https://www.frontiersin.org/journals/digital-health/articles/10.3389/fdgth.2022.1090854/full},
	doi = {10.3389/fdgth.2022.1090854},
	language = {en-US},
	urldate = {2024-12-10},
	journal = {Frontiers in Digital Health},
	author = {Wang, Weinan and Mohseni, Pedram and Kilgore, Kevin L. and Najafizadeh, Laleh},
	month = feb,
	year = {2023},
	note = {Publisher: Frontiers},
}

@article{lee_vitaldb_2022,
	title = {{VitalDB}, a high-fidelity multi-parameter vital signs database in surgical patients},
	volume = {9},
	copyright = {2022 The Author(s)},
	issn = {2052-4463},
	url = {https://www.nature.com/articles/s41597-022-01411-5},
	doi = {10.1038/s41597-022-01411-5},
	language = {en},
	number = {1},
	urldate = {2025-10-23},
	journal = {Scientific Data},
	author = {Lee, Hyung-Chul and Park, Yoonsang and Yoon, Soo Bin and Yang, Seong Mi and Park, Dongnyeok and Jung, Chul-Woo},
	month = jun,
	year = {2022},
	note = {Publisher: Nature Publishing Group},
	pages = {279},
}

@misc{goswami_moment_2024,
	title = {{MOMENT}: {A} {Family} of {Open} {Time}-series {Foundation} {Models}},
	shorttitle = {{MOMENT}},
	url = {http://arxiv.org/abs/2402.03885},
	doi = {10.48550/arXiv.2402.03885},
	urldate = {2025-10-17},
	publisher = {arXiv},
	author = {Goswami, Mononito and Szafer, Konrad and Choudhry, Arjun and Cai, Yifu and Li, Shuo and Dubrawski, Artur},
	month = oct,
	year = {2024},
	note = {arXiv:2402.03885 [cs]},
}

@misc{ansari_chronos_2024,
	title = {Chronos: {Learning} the {Language} of {Time} {Series}},
	shorttitle = {Chronos},
	url = {http://arxiv.org/abs/2403.07815},
	doi = {10.48550/arXiv.2403.07815},
	urldate = {2025-10-17},
	publisher = {arXiv},
	author = {Ansari, Abdul Fatir and Stella, Lorenzo and Turkmen, Caner and Zhang, Xiyuan and Mercado, Pedro and Shen, Huibin and Shchur, Oleksandr and Rangapuram, Syama Sundar and Arango, Sebastian Pineda and Kapoor, Shubham and Zschiegner, Jasper and Maddix, Danielle C. and Wang, Hao and Mahoney, Michael W. and Torkkola, Kari and Wilson, Andrew Gordon and Bohlke-Schneider, Michael and Wang, Yuyang},
	month = nov,
	year = {2024},
	note = {arXiv:2403.07815 [cs]},
}

@article{yun_unsupervised_2024,
	title = {Unsupervised representation learning on high-dimensional clinical data improves genomic discovery and prediction},
	volume = {56},
	copyright = {2024 The Author(s)},
	issn = {1546-1718},
	url = {https://www.nature.com/articles/s41588-024-01831-6},
	doi = {10.1038/s41588-024-01831-6},
	language = {en},
	number = {8},
	urldate = {2025-10-17},
	journal = {Nature Genetics},
	author = {Yun, Taedong and Cosentino, Justin and Behsaz, Babak and McCaw, Zachary R. and Hill, Davin and Luben, Robert and Lai, Dongbing and Bates, John and Yang, Howard and Schwantes-An, Tae-Hwi and Zhou, Yuchen and Khawaja, Anthony P. and Carroll, Andrew and Hobbs, Brian D. and Cho, Michael H. and McLean, Cory Y. and Hormozdiari, Farhad},
	month = aug,
	year = {2024},
	note = {Publisher: Nature Publishing Group},
	pages = {1604--1613},
}

@article{cover1967nearest,
  title={Nearest neighbor pattern classification},
  author={Cover, Thomas and Hart, Peter},
  journal={IEEE transactions on information theory},
  volume={13},
  number={1},
  pages={21--27},
  year={1967},
  publisher={IEEE}
}

@misc{shome_region-disentangled_2023,
	title = {Region-{Disentangled} {Diffusion} {Model} for {High}-{Fidelity} {PPG}-to-{ECG} {Translation}},
	url = {http://arxiv.org/abs/2308.13568},
	language = {british},
	urldate = {2024-11-04},
	publisher = {arXiv},
	author = {Shome, Debaditya and Sarkar, Pritam and Etemad, Ali},
	month = dec,
	year = {2023},
	doi = {10.48550/arXiv.2308.13568},
}

@article{johnson2019billion,
  title={Billion-scale similarity search with GPUs},
  author={Johnson, Jeff and Douze, Matthijs and J{\'e}gou, Herv{\'e}},
  journal={IEEE transactions on big data},
  volume={7},
  number={3},
  pages={535--547},
  year={2019},
  publisher={IEEE}
}

@misc{pillai_papagei_2024,
	title = {{PaPaGei}: {Open} {Foundation} {Models} for {Optical} {Physiological} {Signals}},
	shorttitle = {{PaPaGei}},
	url = {http://arxiv.org/abs/2410.20542},
	urldate = {2025-06-12},
	publisher = {arXiv},
	author = {Pillai, Arvind and Spathis, Dimitris and Kawsar, Fahim and Malekzadeh, Mohammad},
	month = oct,
	year = {2024},
	doi = {10.48550/arXiv.2410.20542},
}

@inproceedings{nambu_cardioflow_2025,
	title = {{CardioFlow}: {Learning} to {Generate} {ECG} from {PPG} with {Rectified} {Flow}},
	shorttitle = {{CardioFlow}},
	url = {https://ieeexplore.ieee.org/document/10888856/},
	doi = {10.1109/ICASSP49660.2025.10888856},
	urldate = {2025-05-19},
	booktitle = {{ICASSP} 2025 - 2025 {IEEE} {International} {Conference} on {Acoustics}, {Speech} and {Signal} {Processing} ({ICASSP})},
	author = {Nambu, Yuta and Kohjima, Masahiro and Yamamoto, Ryuji},
	month = apr,
	year = {2025},
	note = {ISSN: 2379-190X},
	pages = {1--5},
}

@book{dubin2000rapid,
  title={Rapid interpretation of EKG's: an interactive course},
  author={Dubin, Dale},
  year={2000},
  publisher={Cover Publishing Company}
}

@article{castaneda2018review,
  title={A review on wearable photoplethysmography sensors and their potential future applications in health care},
  author={Castaneda, Denisse and Esparza, Aibhlin and Ghamari, Mohammad and Soltanpur, Cinna and Nazeran, Homer},
  journal={International journal of biosensors \& bioelectronics},
  volume={4},
  number={4},
  pages={195},
  year={2018}
}

@article{allen2007photoplethysmography,
  title={Photoplethysmography and its application in clinical physiological measurement},
  author={Allen, John},
  journal={Physiological measurement},
  volume={28},
  number={3},
  pages={R1},
  year={2007},
  publisher={IoP Publishing}
}

@article{garde_development_2014,
	title = {Development of a {Screening} {Tool} for {Sleep} {Disordered} {Breathing} in {Children} {Using} the {Phone} {Oximeter}™},
	volume = {9},
	issn = {1932-6203},
	url = {https://dx.plos.org/10.1371/journal.pone.0112959},
	doi = {10.1371/journal.pone.0112959},
	language = {english},
	number = {11},
	urldate = {2025-09-18},
	journal = {PLoS ONE},
	author = {Garde, Ainara and Dehkordi, Parastoo and Karlen, Walter and Wensley, David and Ansermino, J. Mark and Dumont, Guy A.},
	editor = {Murillo-Rodriguez, Eric},
	month = nov,
	year = {2014},
	pages = {e112959},
}

@article{charlton_2023_2023,
	title = {The 2023 wearable photoplethysmography roadmap},
	issn = {0967-3334},
	url = {http://iopscience.iop.org/article/10.1088/1361-6579/acead2},
	doi = {10.1088/1361-6579/acead2},
	language = {english},
	urldate = {2023-09-27},
	journal = {Physiological Measurement},
	author = {Charlton, Peter H and Allen, John and Bailon, Raquel and Baker, Stephanie and Behar, Joachim A and Chen, Fei and Clifford, Gari D and Clifton, David A and Davies, Harry J and Ding, Cheng and Ding, Xiaorong and Dunn, Jessilyn and Elgendi, Mohamed and Ferdoushi, Munia and Franklin, Daniel and Gil, Eduardo and Hassan, Md Farhad and Hernesniemi, Jussi and Hu, Xiao and Ji, Nan and Khan, Yasser and Kontaxis, Spyridon and Korhonen, Ilkka and Kyriacou, Panayiotis A and Laguna, Pablo and Lazaro, Jesus and Lee, Chungkeun and Levy, Jeremy and Li, Yumin and Liu, Chengyu and Liu, Jing and Lu, Lei and Mandic, Danilo P and Marozas, Vaidotas and Mejía-Mejía, Elisa and Mukkamala, Ramakrishna and Nitzan, Meir and Pereira, Tânia and Poon, Carmen C Y and Ramella-Roman, Jessica C and Saarinen, Harri and Shandhi, Md Mobashir Hasan and Shin, Hangsik and Stansby, Gerard and Tamura, Toshiyo and Vehkaoja, Antti and Wang, Will Ke and Zhang, Yuan-Ting and Zhao, Ni and Zheng, Dingchang and Zhu, Tingting},
	year = {2023},
}

@misc{belhasin_uncertainty-aware_2025,
      title={Uncertainty-Aware PPG-2-ECG for Enhanced Cardiovascular Diagnosis using Diffusion Models}, 
      author={Omer Belhasin and Idan Kligvasser and George Leifman and Regev Cohen and Erin Rainaldi and Li-Fang Cheng and Nishant Verma and Paul Varghese and Ehud Rivlin and Michael Elad},
      year={2025},
      eprint={2405.11566},
      archivePrefix={arXiv},
      primaryClass={cs.LG},
      url={https://arxiv.org/abs/2405.11566}, 
}

\end{document}